\documentclass[letterpaper, 10 pt, conference]{ieeeconf}  
\usepackage{array}
\usepackage{subfigure}
\usepackage{textcomp}
\usepackage{stfloats}
\usepackage{verbatim}
\usepackage{graphicx}
\usepackage{bm}
\usepackage{booktabs}
\usepackage{makecell} 
\usepackage{multirow} 
\usepackage{multicol}
\usepackage{amsmath,amssymb,amsfonts}
\usepackage[ruled]{algorithm2e}  
\usepackage{threeparttable}
\usepackage{cite}
\makeatletter
\let\NAT@parse\undefined
\makeatother
\usepackage[colorlinks,linkcolor=blue,citecolor=black]{hyperref}
\makeatletter
\DeclareRobustCommand\onedot{\futurelet\@let@token\@onedot}
\def\@onedot{\ifx\@let@token.\else.\null\fi\xspace}
 
\def\ie{\emph{i.e}\onedot}

\def\wrt{w.r.t\onedot}

\makeatother

\newcommand{\degcm}[2]{{#1}$^\circ${#2}cm}
\newcommand{\pmdegcm}[2]{$\pm${#1}$^\circ${#2}cm} 
\usepackage[capitalize]{cleveref}
\crefname{section}{Sec.}{Secs.}
\Crefname{section}{Section}{Sections}
\Crefname{table}{Table}{Tables}
\crefname{table}{Tab.}{Tabs.}

\IEEEoverridecommandlockouts                              

\title{\LARGE \bf
Iterative Camera-LiDAR Extrinsic Optimization via Surrogate Diffusion
}

\author{Ni Ou$^{1}$, Zhuo Chen$^{2}$, Xinru Zhang$^{3}$ and Junzheng Wang$^{1}$ 
\thanks{*This work was supported by the National Natural Science Foundation of China under Grant 62173038. The implementation code is available at \href{https://github.com/gitouni/camer-lidar-calib-surrogate-diffusion}{https://github.com/gitouni/camer-lidar-calib-surrogate-diffusion}.}
\thanks{$^{1}$Ni Ou and Junzheng Wang are with the School of Automation, Beijing Institute of Technology, Beijing, 100081, China.
        {\tt\small wangjz@bit.edu.cn}}%
\thanks{$^{2}$Zhuo Chen is with the Robot Perception Lab, Centre for Robotics Research, Department of Engineering, King's College London, London WC2R 2LS, United Kingdom.}%
\thanks{$^{3}$Xinru Zhang is with the School of Integrated Circuits and Electronics, Beijing Institute of Technology, Beijing, 100081, China.}%
}

\begin{document}

\maketitle
\thispagestyle{empty}
\pagestyle{empty}

\begin{abstract}
Cameras and LiDAR are essential sensors for autonomous vehicles. The fusion of camera and LiDAR data addresses the limitations of individual sensors but relies on precise extrinsic calibration. Recently, numerous end-to-end calibration methods have been proposed; however, most predict extrinsic parameters in a single step and lack iterative optimization capabilities. To address the increasing demand for higher accuracy, we propose a versatile iterative framework based on surrogate diffusion. This framework can enhance the performance of any calibration method without requiring architectural modifications. Specifically, the initial extrinsic parameters undergo iterative refinement through a denoising process, in which the original calibration method serves as a surrogate denoiser to estimate the final extrinsics at each step. For comparative analysis, we selected four state-of-the-art calibration methods as surrogate denoisers and compared the results of our diffusion process with those of two other iterative approaches. Extensive experiments demonstrate that when integrated with our diffusion model, all calibration methods achieve higher accuracy, improved robustness, and greater stability compared to other iterative techniques and their single-step counterparts.
\end{abstract}

\section{INTRODUCTION}
Camera and LiDAR are two of the most popular sensors applied in autonomous driving. The camera captures colorful images with dense semantic context, while the LiDAR measures distances of sparse points with intensity that reflect the rough outline of the ambient scene. Their data fusion compensates the limitations of stand-alone sensors and has been involved in a large variety of downstream intelligent transportation tasks, such as 3D object detection~\cite{Object-Detection1,Oject-Detection2},  simultaneously localization and mapping (SLAM)~\cite{SLAM1,SLAM2} and scene flow estimation~\cite{SceneFlow1,CamLiFlow}. 

Camera-LiDAR calibration is the prerequisite for the aforementioned tasks, since it establishes the spatial relationship between the two sensors. The evolution of deep learning techniques has significantly advanced the development of learning-based calibration methods~\cite{CalibNet,RGGNet,LCCNet,LCCRAFT,DeepI2P,ATOP}. These methods either explicitly or implicitly identify correspondences between image and point cloud features to predict the corrections to the extrinsic parameters. Yet, most of these approaches produce calibration results in a single step, thereby leaving subsequent states after the initial adjustment unexploited. This oversight may limit the final accuracy because further refinements could improve accuracy, especially when the initial error is substantial.

To address this issue, CalibNet~\cite{CalibNet} introduces a straightforward single-model iterative approach: for each iteration, the output of the surrogate is used to correct the input extrinsics, forming the input of the next iteration. However, the success of this iterative process heavily relies on the original model's capability and robustness, specifically its ability to enhance accuracy across a wide range of initial errors. Multi-range iteration~\cite{LCCNet} alleviates this issue by training different models for various error ranges. Each model is tasked with reducing the calibration error to the next lower level, allowing the entire system to incrementally minimize error to the lowest possible range. Despite success in improving calibration accuracy, it necessitates separate training, inference, and storage for each model. This need for additional memory and computational resources presents challenges for online calibration, particularly when deploying on edge-computing devices in autonomous vehicles.

\begin{figure}[!t]
    \centering
    \includegraphics[width=0.98\linewidth]{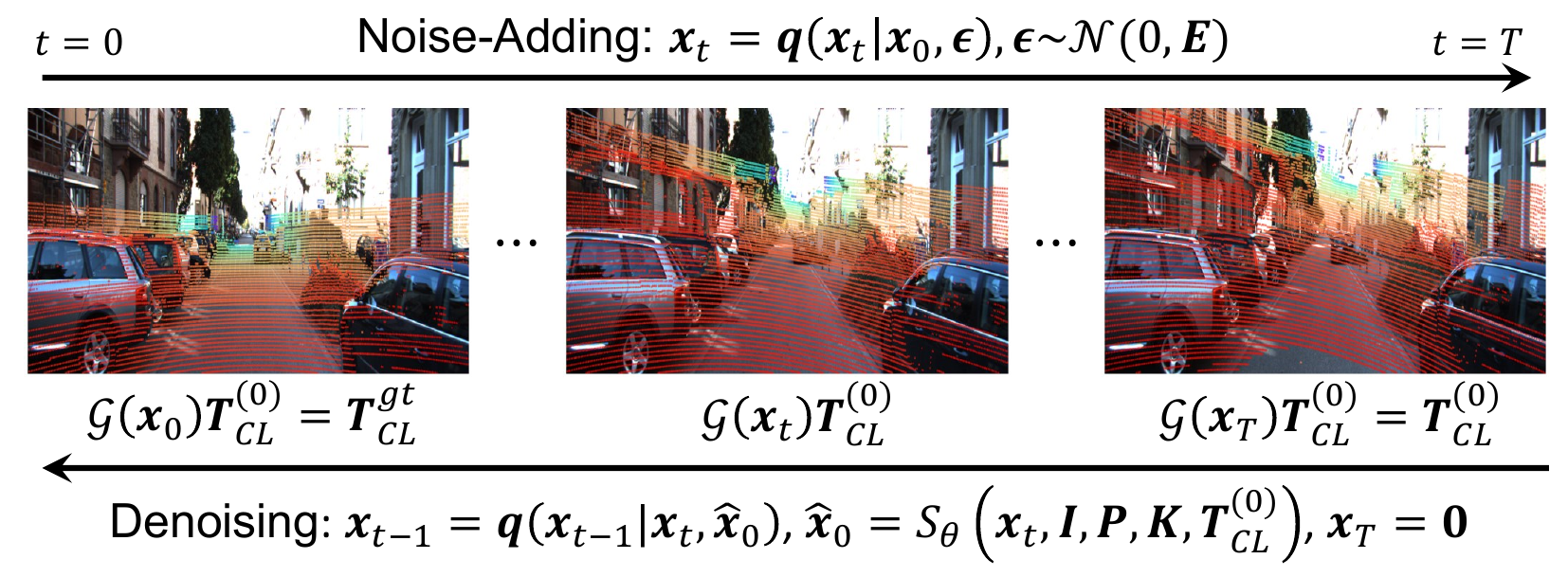}
    \caption{The proposed surrogate diffusion for camera-LiDAR calibration. The diffusion variable $\bm{x}_t$ controls the correction factor $\mathcal{G}(\bm{x}_t)$ applied to the initial extrinsic matrix $\bm{T}_{CL}^{(0)}$ to generate noisy samples $\mathcal{G}(x_t)\bm{T}_{CL}^{(0)}$. The noise-adding process transfers a ground-truth extrinsic matrix to an initial state that contains Gaussian noise, while the denoising process reverses it by applying a trainable surrogate $S_\theta$.}
    \label{Fig.abstract}
\end{figure}

In this study, we propose an innovative single-model iterative method that can improve any surrogate model through diffusion. To the best of our knowledge, this is the first application of diffusion in the context of camera-LiDAR calibration. As illustrated in \cref{Fig.abstract} (with notations defined in \cref{sec.method}), the original method serves as a surrogate to iteratively refine the initial extrinsic matrix until it converges to the ground-truth matrix. The main contributions of our paper are outlined below:

\begin{itemize}
\item A linear surrogate diffusion (LSD) pipeline is proposed for single-model iterative camera-LiDAR calibration optimization. It is denoiser-agnostic and applicable to any individual calibration method.
\item We analyze the data flow of our iterative approach and develop an intermediate buffer to enhance efficiency during the reverse LSD process.
\item Extensive experiments on the KITTI dataset~\cite{KITTI} have been conducted to validate the effectiveness and efficiency of our proposed diffusion method.
\end{itemize}

The remainder of this paper is organized as follows. \Cref{sec.related_work} reviews recent target-based and targetless calibration methods; \Cref{sec.method} introduces the pipeline of our surrogate diffusion model; \Cref{sec.experiments} presents the experimental settings and results; \Cref{sec.conclusion} summarizes our findings and discusses our future study.

\section{RELATED WORKS}
\label{sec.related_work}
\subsection{Target-Based Calibration Methods}
Target-based calibration determines the extrinsic matrix between camera and LiDAR by utilizing a specific target that incorporates geometric constraints between corresponding 3D points in the point cloud and pixels in the 2D image. Calibration targets are classified into planar and 3D objects based on their shapes. Planar targets include chessboards~\cite{Calib-line-plane,Calib-geometry,Calib-checkboard-passthrough}, triangular boards~\cite{Calib-triangular, Calib-triangular2} and boards with holes~\cite{Calib-Rectangle-Hole,Calib-Cycle-Hole,Calib-cycle-hole2}. In contrast, 3D calibration tools comprise V-shaped~\cite{Calib-Vshape} and box-shaped objects~\cite{Calib-Box}. Despite high accuracy and reproducibility, target-based calibration methods encounter several challenges, including the requirement for manual target placement in diverse positions and limited suitability for online calibration. Furthermore, determining certain hyperparameters, such as target size and calibration distance, remains challenging across different sensor systems.

\subsection{Targetless Calibration Methods}
Instead of relying on the introduction of specific calibration targets, targetless methods leverage information extracted from natural scenes for calibration. These methods can be broadly categorized into four groups~\cite{Calib-Survey}: ego-motion-based, feature-based, information-based, and learning-based. Ego-motion-based methods hinges on geometric constraints spanning multiple frames, exemplified by techniques like hand-eye calibration~\cite{Calib-HE1,Calib-HE2} and modality-consistent 3D reconstruction~\cite{Calib-SfM1,Calib-SfM2,Calib-CMSC}. Feature-based methods solve extrinsics through cross-modal feature extraction and matching, using hand-crafted features such as edge points~\cite{Calib-edge1,Calib-edge2,Calib-edge3} and planar constraints~\cite{Calib-plane-constrain}, or mask matching based on semantic information~\cite{Calib-SAM-Align, Calib-Cross-Mask-Matching,Calib-SAM-Lightweight}. Information-based methods optimize an information metric like mutual information~\cite{Calib-MI1,Calib-MI2} or normalized mutual information~\cite{Calib-NMI1,Calib-NMI2}. Learning-based methods learn cross-modal correspondences~\cite{CorrI2P,Voxelpoint-to-pixel-matching,CoFiI2P} or employ a end-to-end calibration network~\cite{CalibNet,LCCNet,RGGNet,CalibDepth,LCCRAFT}.

\subsection{End-to-End Learning-based methods}
End-to-end learning-based methods are central to our research. CalibNet \cite{CalibNet} exemplifies a typical end-to-end calibration network, using ResNet \cite{Resnet} to extract features from camera and LiDAR data, which are then fused via convolutional and MLP layers. Building upon this framework, RGGNet introduces a regularization loss to guide the network in predicting extrinsics that align with the ground-truth data distribution. LCCNet \cite{LCCNet} enhances accuracy with a feature-matching layer that explicitly aligns deep features of images and point clouds, employing multi-range iterations. LCCRAFT \cite{LCCRAFT} simplifies the encoders of LCCNet \cite{LCCNet} and utilizes a RAFT-like \cite{RAFT} architecture for iterative and alternating optimization of extrinsic and feature matching predictions. CalibDepth \cite{CalibDepth} utilizes monocular depth maps to enhance cross-modality feature matching and implements LSTM for multi-step prediction.

In our experiments, we selected CalibNet, RGGNet, LCCNet, and LCCRAFT as surrogate denoisers due to their identical input modalities. To validate the effectiveness of our iterative approach, we combined these models with various iterative techniques to assess performance improvements. We selected two additional single-model iterative approaches as baselines: the straightforward iterative method proposed in \cite{CalibNet} and SE(3) Diffusion~\cite{SE3-Diffusion}, which was originally developed for point cloud registration and is related to our LSD. We adapted SE(3) Diffusion for camera-LiDAR calibration to enable a comparative analysis.

\section{METHOD}
\label{sec.method}
\subsection{Problem Setting}
\label{subsec.problem_setting}
Let $\bm{I}$ represent the RGB image captured by the camera and $\bm{P}$ denote the LiDAR point cloud. Define the relative transformation from LiDAR to camera as $\bm{T}_{CL}\in \textup{SE(3)}$ and the intrinsic matrix of the camera as $\bm{K}$. Suppose that we have known $\bm{K}$ and an initial guess of $\bm{T}_{CL}^{gt}$, denoted as $\bm{T}_{CL}^{(0)}$. For simplicity, we use $\bm{C}$ to represent the conditions $[\bm{I},\bm{P},\bm{K}]$. Given $\bm{C}$ and $\bm{T}_{CL}^{(0)}$, the objective of a camera-LiDAR calibration method $D_\theta$ is to estimate $\bm{T}_{CL}^{gt}$. Since we have known the initial extrinsic matrix $\bm{T}_{CL}^{(0)}$, we expect $D_\theta$ to output the correction to the left transformation, \ie, $\bm{T}_{CL}^{gt}(\bm{T}_{CL}^{(0)})^{-1}$. Considering the internal constraints on parameters of this SE(3) matrix are challenging for neural networks to process, we convert it to the Lie algebra form as the desired output of $D_\theta$:
\begin{equation}
    \label{Eq.denoiser_io}
    \Delta \bm{\xi}_{gt}=\mathcal{G}^{-1}\left(\bm{T}_{CL}^{gt}(\bm{T}_{CL}^{(0)})^{-1}\right) \in \mathfrak{se}(3)
\end{equation}
where $\mathcal{G}$ is the exponential map from $\mathfrak{se}(3)$ to SE(3), and $\mathcal{G}^{-1}$ is its inverse function.

The loss function to supervise $D_\theta$ is:
\begin{equation}
    \label{Eq.denoiser_loss}
    \mathcal{L}(\Delta \hat{\bm{\xi}}_{gt},  \Delta \bm{\xi}_{gt}) = \Vert \Delta \hat{\bm{\xi}}_{gt}  - \Delta \bm{\xi}_{gt} \Vert_1
\end{equation}
where $\Delta \hat{\bm{\xi}}_{gt}$ denotes the output of $D_\theta$.

To obtain the final estimation for $\bm{T}_{CL}^{gt}$, we just need to left multiply the SE(3) output of $D_\theta$ to $\bm{T}_{CL}^{(0)}$ as follows:
\begin{equation}
    \label{Eq.denoiser_surrogate}
    \hat{\bm{T}}_{CL}^{gt}=\mathcal{G}(\Delta\hat{\bm{\xi}}_{gt})\bm{T}_{CL}^{(0)} = \mathcal{G}\left(D_\theta(\bm{C},\bm{T}_{CL}^{(0)})\right)\bm{T}_{CL}^{(0)}
\end{equation}

To extend the above single-step prediction into a naive iterative method (NaIter), the current output can be utilized as the input for the subsequent iteration:
\begin{equation}
    \label{Eq.naive_iteration}
    \begin{cases}
    \hat{\bm{T}}_{CL}^{(i)} = \Delta\hat{\bm{T}}_{CL}^{(i)}\bm{T}_{CL}^{(0)}\; ,\Delta \hat{\bm{T}}_{CL}^{(0)}=\bm{E} \\
    \Delta\hat{\bm{T}}_{CL}^{(i+1)} = \mathcal{G}\left(D_\theta(\bm{C},\hat{\bm{T}}_{CL}^{(i)})\right)\Delta\hat{\bm{T}}_{CL}^{(i)}
    \end{cases}
\end{equation}

\subsection{Linear Surrogate Diffusion}
\label{subsec.diffusion}
\subsubsection{Review of Diffusion Models}
\label{subsubsec.review_diffusion}
Diffusion models~\cite{DDPM,DPM,UNIPC} is a category of likelihood-based generative models including a forward and reverse process. During the forward process $\bm{q}(\bm{x}_t|\bm{x}_{t-1})$, noise is progressively added to the sample $\bm{x}_0$ to generate noisy sample $\bm{x}_t$ until transforming it into pure Gaussian noise $\bm{\epsilon}\sim \mathcal{N}(\bm{0},\bm{E})$ ($\bm{E}$ is an identical matrix). This process can be simplified into a close form expression $\bm{q}(\bm{x}_t|\bm{x}_0,\bm{\epsilon})$:
\begin{equation}
    \label{Eq.definition_x_t}
    \bm{x}_t = \bm{q}(\bm{x}_t|\bm{x}_0,\bm{\epsilon})=\sqrt{\overline{\alpha}_t}\bm{x}_{0} + \sqrt{1-\overline{\alpha}_t}\bm{\epsilon}
\end{equation}
where $\overline{\alpha}_t$ is subject to a certain noise schedule. Here we adopt the cosine noise schedule proposed in~\cite{Improved-DDPM}, as formulated in \cref{Eq.noise_schedule}.
\begin{equation}
\label{Eq.noise_schedule}
    \begin{cases}
    \overline{\alpha}_t=\frac{f(t)}{f(0)},\,f(t)=cos\left(\frac{t/T+s}{1+s}\cdot \frac{\pi}{2}\right)^2\\
    \alpha_t = 1 - \beta_t,\,\beta_t = 1 - \frac{\overline{\alpha}_t}{\overline{\alpha}_{t-1}}
\end{cases}
\end{equation}

Assume that the learned network estimates $\bm{x}_0$ as $\hat{\bm{x}}_0$. The reverse process aims to establish a probability $\bm{q}(\bm{x}_{t-1}|\bm{x}_{t},\hat{\bm{x}}_0)$, iteratively recovering $\bm{x}_0$ from $\bm{x}_T$. The standard denoising probability diffusion model~\cite{DDPM} utilizes a stochastic reverse process formulated as:
\begin{equation}
    \label{Eq.diffusion_x_t_1}
    \bm{x}_{t-1}=\bm{q}(\bm{x}_{t-1}|\bm{x}_{t},\hat{\bm{x}}_0)=\bm{\mu}_\theta(\bm{x}_t,\hat{\bm{x}}_0,t) + \bm{\Sigma}(t)\bm{\epsilon}
\end{equation}
where $\bm{\mu}_\theta(\bm{x}_t,\hat{\bm{x}}_0,t)$ and $\bm{\Sigma}(t)$ are formulated as:
\begin{equation}
\label{Eq.mean_x_t}
    \bm{\mu}_\theta(\bm{x}_t,\hat{\bm{x}}_0,t)=\frac{\sqrt{\alpha_t}(1-\overline{\alpha}_{t-1})\bm{x}_t+\sqrt{\overline{\alpha}_{t-1}}(1-\alpha_t)\hat{\bm{x}}_0}{1-\overline{\alpha}_t}
\end{equation}
\begin{equation}
    \label{Eq.sigma_x_t}
    \bm{\Sigma}(t)=\frac{(1-\alpha_t)(1-\overline{\alpha}_{t-1})}{1-\overline{\alpha}_t}\bm{E}
\end{equation}
\subsubsection{Selection of the Diffusion Variable}
\label{subsubsec.diffusion_variable_selection}
\begin{algorithm}[!t]
	\caption{Diffusion Process (for training)}
	\label{algo.DP}
	\KwIn{$\bm{T}_{CL}^{gt}, \bm{T}_{CL}^{(0)}, \{\overline{\alpha}_t\}_{i=1}^T, \bm{I}, \bm{P}, \bm{K}, N$}
	$\bm{x}_0=\mathcal{G}^{-1}(\bm{T}_{CL}^{gt}(\bm{T}_{CL}^{(0)})^{-1})$\\
        $\bm{\epsilon}=\bm{0}$\\
	\For{$i = 1,2,...,N$}
	{
            Randomly select $t$ from $\{1,2,...,T\}$ \\
		$\bm{x}_t = \sqrt{\overline{\alpha}_t}\bm{x}_{0} + \sqrt{1-\overline{\alpha}_t}\bm{\bm{\epsilon}_}0$\\
            Compute $\hat{\bm{x}}_0$ using \cref{Eq.diffusion_x0} \\
            Compute loss $\mathcal{L}_{LSD}$ using \cref{Eq.diffusion_loss}\\
            Backpropagate the gradient \wrt $\theta$\\
	}
\end{algorithm}
As shown in~\cref{Fig.abstract}, unlike diffusion models for image generation~\cite{DDPM,DDIM,DPM}, a diffusion model for camera-LiDAR calibration requires denoising on the extrinsic matrix $\bm{T}_{CL}$, which contains internal SE(3) constraints. Another difference is that the initial state of our diffusion should be centered around the initial extrinsic matrix $\bm{T}_{CL}^{(0)}$ rather than pure Gaussian noise.

Based on the above analysis, we model our diffusion process on the transformation difference between $\bm{T}_{CL}^{gt}$ and $\bm{T}_{CL}^{(0)}$ and retrieve its Lie algebra form as our variable. In this case, the noisy initial extrinsic matrix can be expressed as $\mathcal{G}(\bm{x}_t)\bm{T}_{CL}^{(0)}$. As for the boundary constraints, $\bm{x}_T$ is set to $\bm{0}$ to ensure $\mathcal{G}(\bm{x}_T)\bm{T}_{CL}^{(0)}=\bm{T}_{CL}^{(0)}$, and $\bm{x}_0$ is set to $\Delta \bm{\xi}_{gt}$ (defined in~\cref{Eq.denoiser_io}) to satisfy $\mathcal{G}(\bm{x}_0)\bm{T}_{CL}^{(0)}=\bm{T}_{CL}^{gt}$.

This definition results in $\bm{\epsilon} = \bm{x}_T = \bm{0}$, suggesting that $\bm{\epsilon}$ follows a Dirac Distribution $\delta(\bm{0})$. Although this setting may appear counterintuitive, we can regard it as a general diffusion process defined in~\cite{Cold-Difffusion}. Additionally, the condition $\bm{\epsilon}\neq \bm{0}$ increases the variation of $\Delta \bm{\xi}_{gt}$, which will be adverse to the inverse process.  Therefore, we decide to retain the setting of $\bm{\epsilon} = \bm{x}_T = \bm{0}$.

\subsubsection{Surrogate Formulation}
\label{surrogate_formulation}
Inspired by~\cite{SE3-Diffusion}, we introduce a surrogate to make our diffusion denoiser-agnostic. The surrogate $S_\theta$ omits the time embedding layer and estimates the transformation difference between $\bm{T}_{CL}^{(0)}$ and $\bm{T}_{CL}^{gt}$ from the noisy input $\bm{x}_t$, which can be mathematically expressed as $\hat{\bm{x}}_0=S_\theta(\bm{x}_t,\bm{C},\bm{T}_{CL}^{(0)})=\mathcal{G}^{-1}(\hat{\bm{T}}_{CL}^{gt}(\bm{T}_{CL}^{(0)})^{-1})$. As described in~\cref{subsec.problem_setting}, $D_\theta$ predicts the transformation difference between $\bm{T}_{CL}^{gt}$ and $\bm{T}_{CL}^{(0)}$. Therefore, the relationship of $D_\theta$ and $\hat{\bm{x}}_0$ can be formulated as:
\begin{equation}
    \label{Eq.D_to_S}
    \underbrace{\mathcal{G}(\hat{\bm{x}}_0)\bm{T}_{CL}^{(0)}}_{\hat{\bm{T}}_{CL}^{gt}}=\underbrace{\mathcal{G}\left(D_\theta(\bm{C},\mathcal{G}(\bm{x}_t)\bm{T}_{CL}^{(0)})\right)}_{D_\theta \;\textbf{output}}\underbrace{\mathcal{G}(\bm{x}_t)\bm{T}_{CL}^{(0)}}_{D_\theta\; \textbf{input}}
\end{equation}
which can be simplified as below:
\begin{equation}
    \label{Eq.diffusion_x0}
   \hat{\bm{x}}_0 = \mathcal{G}^{-1}\left(\mathcal{G}\left(D_\theta(\bm{C},\mathcal{G}(\bm{x}_t)\bm{T}_{CL}^{(0)})\right)\mathcal{G}(\bm{x}_t)\right)
\end{equation}
In this context, the loss function to supervise $D_\theta$ is:
\begin{equation}
    \label{Eq.diffusion_loss}
    \mathcal{L}_{LSD}(\hat{\bm{x}}_0,\bm{x}_0) = \Vert \hat{\bm{x}}_0 - \bm{x}_0\Vert_1
\end{equation}

\begin{algorithm}[!t]
	\caption{Reverse Process (for inference)}
	\label{algo.RP}
	\KwIn{$\bm{T}_{CL}^{(0)}, \{\alpha_t\}, \{\overline{\alpha}_t\}_{i=1}^T, \bm{I}, \bm{P}, \bm{K}$}
	\KwOut{$\hat{\bm{T}}_{CL}^{gt}$}
    $\bm{x}_T=\bm{\epsilon}=\bm{0}$\\
	\For{$t=T,T-1,...,1$}
	{
Compute $\hat{\bm{x}}_0$ using \cref{Eq.diffusion_x0} \\
Compute $\bm{x}_{t-1}=\bm{q}(\bm{x}_{t-1}|\bm{x}_{t},\hat{\bm{x}}_0)$ using~\cref{Eq.diffusion_x_t_1} \\
	}
\Return $\hat{\bm{T}}_{CL}^{gt}=\mathcal{G}(\bm{x}_0)\bm{T}_{CL}^{(0)}$
\end{algorithm}
In summary, during the forward process, $\bm{x}_t$ is obtained using~\cref{Eq.definition_x_t} and serves as the input of the $S_\theta$, while $D_\theta$ is supervised by \cref{Eq.diffusion_loss}. The entire forward process is summarized in \cref{algo.DP}. Concerning the reverse process, $\bm{x}_T$ is initialized as $\bm{0}$ and progressively recovered into $\bm{x}_0$ applying~\cref{Eq.diffusion_x0} and ~\cref{Eq.diffusion_x_t_1} alternately. The whole reverse process is outlined in Algorithm~\ref{algo.RP}. For clarity, we take DDPM~\cite{DDPM} as an example to introduce our reverse process, but its sampler can be replaced with other efficient ODE solvers such as DPM~\cite{DPM} and UniPC~\cite{UNIPC}.

\subsubsection{Intermediate Variable Buffering}
\label{subsubsec.buffering}
Regarding the proposed surrogate model, the initial extrinsic matrix varies with $t$ according to \cref{Eq.diffusion_x0}. However, we observe that some intermediate variables remain unchanged from the second iteration so that they can be stored in the first iteration for subsequent reusing. For example, the common operation of CalibNet, RGGNet, LCCNet and LCCRAFT is the image feature extraction, which is independent from $\bm{T}_{CL}$, thus the extracted image feature can be reused after the first iteration. Intermediate variable buffering is implemented during inference. Specifically, in Algorithm~\ref{algo.RP}, it should be employed when $t=T-1,...,1$. We found this modification is also applicable to other iterative techniques and apply it to all of them for fair efficiency comparison.
\section{EXPERIMENTS}
\label{sec.experiments}
\subsection{Dataset Description}
\label{subsec.dataset}
We conduct calibration experiments on the KITTI Odometry Dataset~\cite{KITTI} that contains 22 sequences of camera-LiDAR data with corresponding ground-truth extrinsic matrices $\bm{T}_{CL}^{gt}$ and intrinsic matrices $\bm{K}$. To generate initial transformations $\bm{T}_{CL}^{(0)}$ for the inputs, random perturbations are imposed on $\bm{T}_{CL}^{gt}$, of which the rotation and translation ranges are respectively set to $\pm15^\circ$ and $\pm 15$cm on each axis (referred to as \pmdegcm{15}{15} hereinafter). For the data division, sequences 00, 02, 03, 04, 05, 06, 07, 08, 10, 12 are chosen for training, sequences 16, 17, 18 for validation, and sequences 13, 14, 15, 20, 21 for testing.
\subsection{Implementation Details}
\label{subsec.implementation}
The image encoders of CalibNet, RGGNet and LCCNet are all configured to ResNet-18~\cite{Resnet}. Since the public code of LCCRAFT is unavailable, we implemented its image encoder using the default hyperparameters of RAFT~\cite{RAFT}.

Regarding diffusion settings, $s$ is set to 0.008 in \cref{Eq.noise_schedule} for our noise schedule. We use the LogSNR sampling scheduler and apply the UniPC~\cite{UNIPC} sampler to replace DDPM in \cref{algo.RP} for acceleration. The number of function evaluations (NFE) for all iterative methods is set to 10.

To demonstrate the advantages of our \textbf{LSD} approach, we compare it with single-use (\textbf{Single}) defined in \cref{Eq.denoiser_surrogate} and \textbf{NaIter} formulated in \cref{Eq.naive_iteration}. Additionally, we adapt a surrogate diffusion model, originally used in point cloud registration, to this calibration task for comparative purposes. We refer to this model as non-linear surrogate diffusion (\textbf{NLSD}). The differences among these iterative methods are discussed in \cref{subsubsec.diff_iterative}.

\subsection{Metrics}
\label{subsec.metrics}
We apply several metrics to comprehensively evaluate the performance of our method and baselines. These metrics are defined based on the SE(3) distance:
\begin{equation}
    \label{Eq.se3_dist}
    \bm{\varepsilon_T} = \hat{\bm{T}}_{CL}^{gt}(\bm{T}_{CL}^{gt})^{-1} \in \textup{SE(3)}
\end{equation}

To qualify calibration accuracy, we record the Euler angles of each axis (\textbf{Rx, Ry, Rz}) and translation values of each axis (\textbf{tx, ty, tz}) \wrt $\bm{\varepsilon_T}$, together with rotation and translation root squared mean error (\textbf{RMSE}).

To evaluate calibration robustness, another two metrics are designed to illustrate the proportion of valid samples on which the calibration errors are within a certain range. Specifically, the metric \textbf{\degcm{3}{3}} reflects the percentage of samples with rotation and translation RMSE under 3$^\circ$ and $3$cm respectively, and a similar definition applies to \textbf{\degcm{5}{5}}.

Additionally, we evaluated the stability of different iterative methods, which is defined by the degree of monotonic decrease in iteration count and accuracy. Similar to \degcm{3}{3}, a metric named $\bm{\rho}$\textbf{\%} is designed to measure the proportion of samples whose rotation RMSE and translation RMSE both satisfy the following equation:
\begin{equation}
    \textup{RMSE}_{i=2} \geq \textup{RMSE}_{i=5} \geq \textup{RMSE}_{i=10}
\end{equation}
, where $\textup{RMSE}_{i=k}$ represents the rotation/translation RMSE of the $k^\textup{th}$ iteration. The above equation reflects a property where the more iterations undergoes, the higher accuracy achieved by the model.

\subsection{Calibration Results}
\label{subsec.exp_iterative}
\subsubsection{Calibration Accuracy}
\label{subsubsec.calib_acc}
\begin{figure}[!t]
    \centering
    \subfigure[Rotation RMSE ($^\circ$){\label{Fig.RRMSE_boxplot}}]{\includegraphics[width=0.95\linewidth]{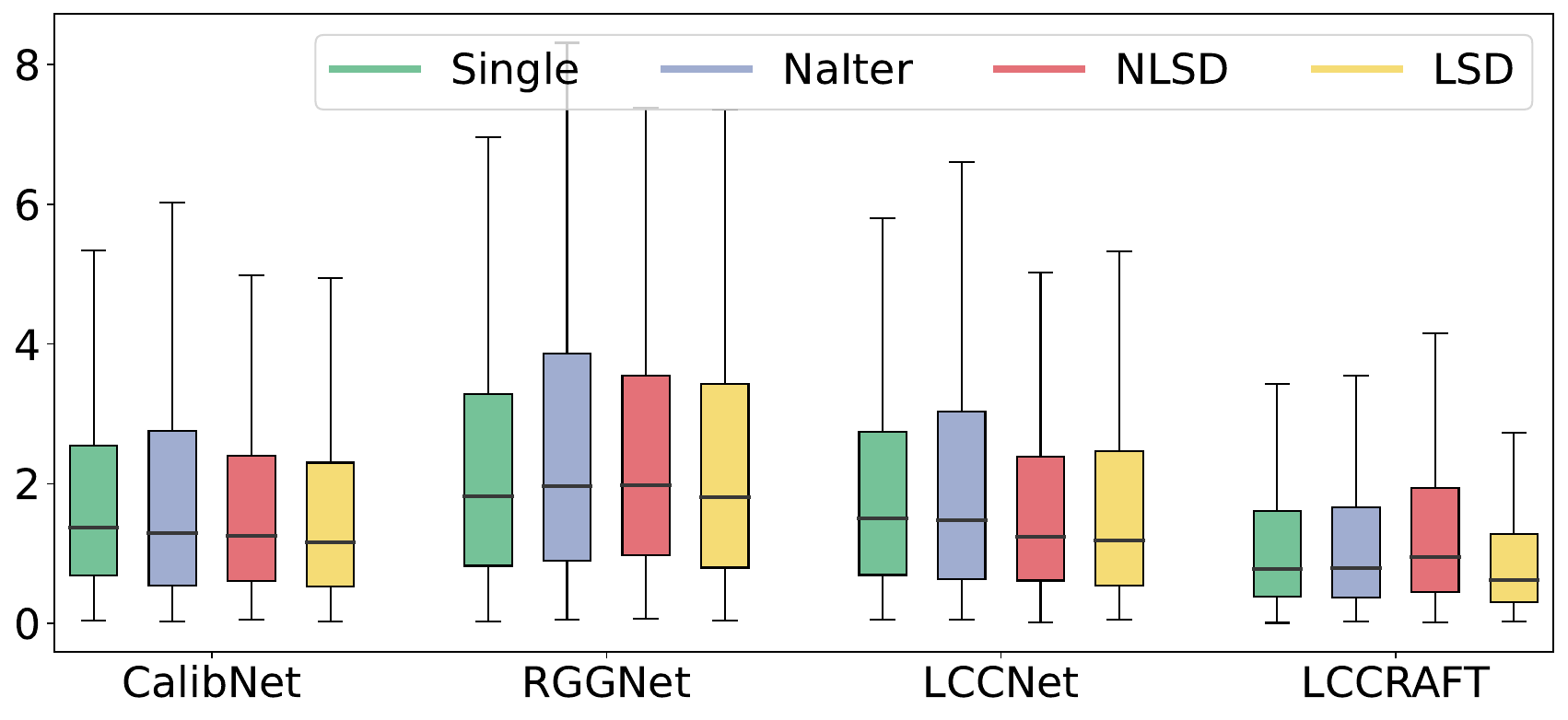}}
		
    \subfigure[Translation RMSE (cm){\label{Fig.TRMSE_boxplot}}]{\includegraphics[width=0.95\linewidth]{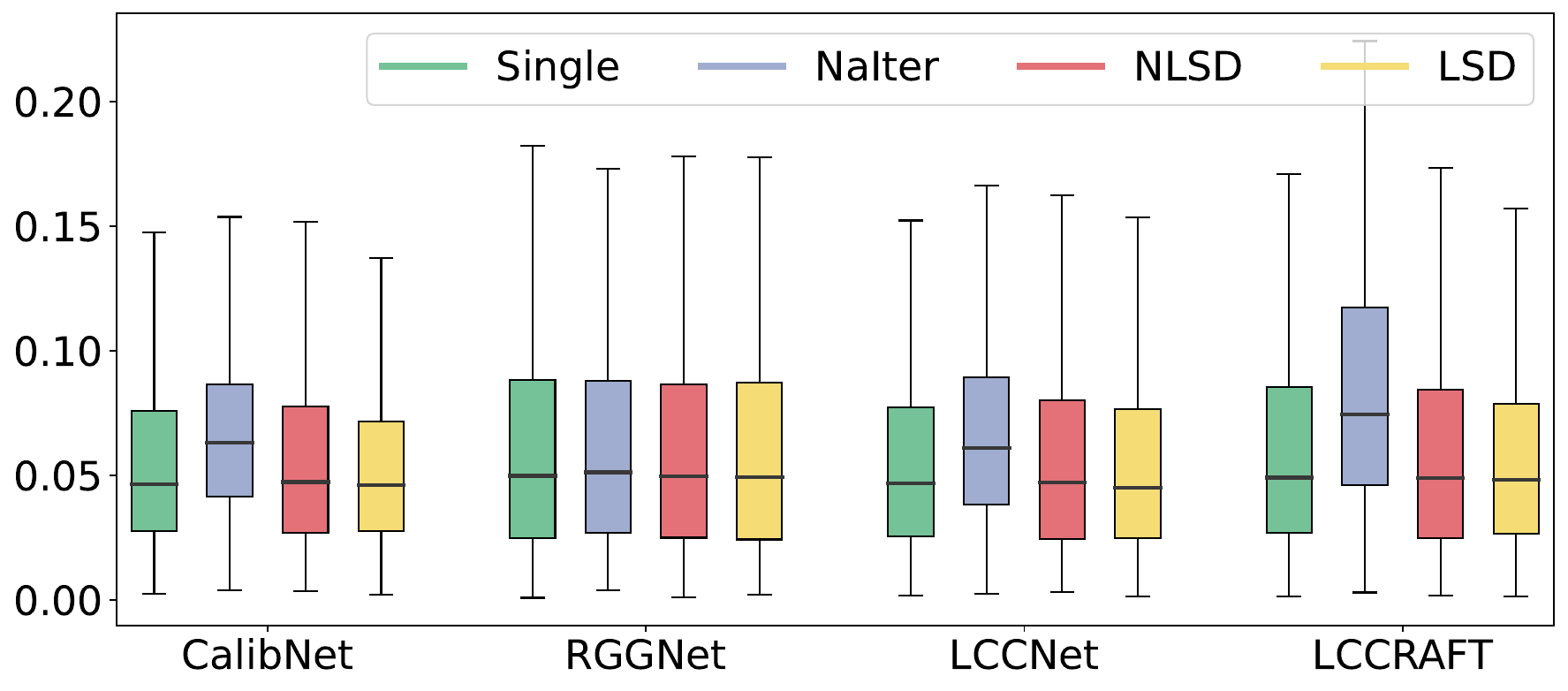}}
    \caption{Distribution of Rotation RMSE and Translation RMSE of Different Iterative Methods.}
    \label{Fig.boxplot}
\end{figure}
\Cref{Fig.boxplot} illustrates the distribution of rotation and translation RMSE for Single, NaIter, NLSD, and LSD. For rotation RMSE, LSD consistently outperforms the other iterative methods across all surrogates. NaIter exhibits the poorest performance and the largest variation in most cases, except for CalibNet. NLSD does not consistently outperform Single across all surrogates. It performs better than Single in CalibNet and LCCNet but underperforms in RGGNet and LCCRAFT.

In terms of translation RMSE, LSD demonstrates superior performance in LCCNet and LCCRAFT, though its advantage over Single is not as pronounced as in rotation RMSE. The median errors and variations for NLSD are higher compared to LSD. NaIter again performs the worst across all surrogates, although its variation is close to those of other iterative methods.
\subsubsection{Calibration Robustness and Stability}
\label{subsubsec.calib_robust_stable}
\begin{table}[htbp]
\centering
    \caption{Calibration Robustness and Stability}
    \label{Table.Calib-Error-Iterative}
        \begin{tabular}{cccc}  
            \toprule
            Method &\degcm{3}{3}$\uparrow$ &\degcm{5}{5}$\uparrow$ &$\rho$\%$\uparrow$ \\
        \midrule
CalibNet (Single)~\cite{CalibNet} &23.19\% &49.37\% &N/A \\
RGGNet (Single)~\cite{RGGNet} &22.04\% &43.53\% &N/A \\
LCCNet (Single)~\cite{LCCNet} &23.88\% &48.47\% &N/A \\
LCCRAFT (Single)~\cite{LCCRAFT} &26.38\% &47.33\% &N/A \\
\midrule
CalibNet + NaIter &12.50\% &32.75\% &2.98\% \\
RGGNet + NaIter &19.65\% &39.90\% &8.55\% \\
LCCNet + NaIter &13.28\% &34.58\% &4.74\% \\
LCCRAFT + NaIter &10.39\% &27.45\% &4.75\% \\
\midrule
CalibNet + NLSD &23.46\% &47.96\% &7.66\% \\
RGGNet + NLSD &20.67\% &43.04\% &6.19\% \\
LCCNet + NLSD &26.15\% &48.94\% &7.15\% \\
LCCRAFT + NLSD &26.29\% &46.74\% &7.16\% \\
\midrule
CalibNet + LSD &\textbf{24.39\%} &\textbf{49.52\%} &\textbf{38.62\%} \\
RGGNet + LSD &\textbf{22.24\%} &\textbf{44.09\%} &\textbf{38.86\%} \\
LCCNet + LSD &\textbf{26.27\%} &\textbf{50.14\%} &\textbf{45.54\%} \\
LCCRAFT + LSD &\textbf{27.90\%} &\textbf{49.96\%} &\textbf{47.61\%} \\
\bottomrule
\end{tabular}
\end{table}

On top of accuracy, we also compare the robustness and stability of these iterative methods in \cref{Table.Calib-Error-Iterative}. The results indicate that LSD surpasses the other two iterative methods across all three metrics, with a particularly significant advantage in terms of $\rho$\%. In contrast, NaIter is the most unstable iterative method and lacks robustness. While NLSD exhibits improved robustness over Single on CalibNet and LCCNet, it does not show similar improvements on the other two surrogates. Furthermore, the $\rho$\% of NLSD remains notably inferior to that of LSD.

\subsection{Differences of Three Iterative Methods}
\label{subsubsec.diff_iterative}
\begin{figure*}[!t]
    \centering
    \includegraphics[width=0.95\linewidth]{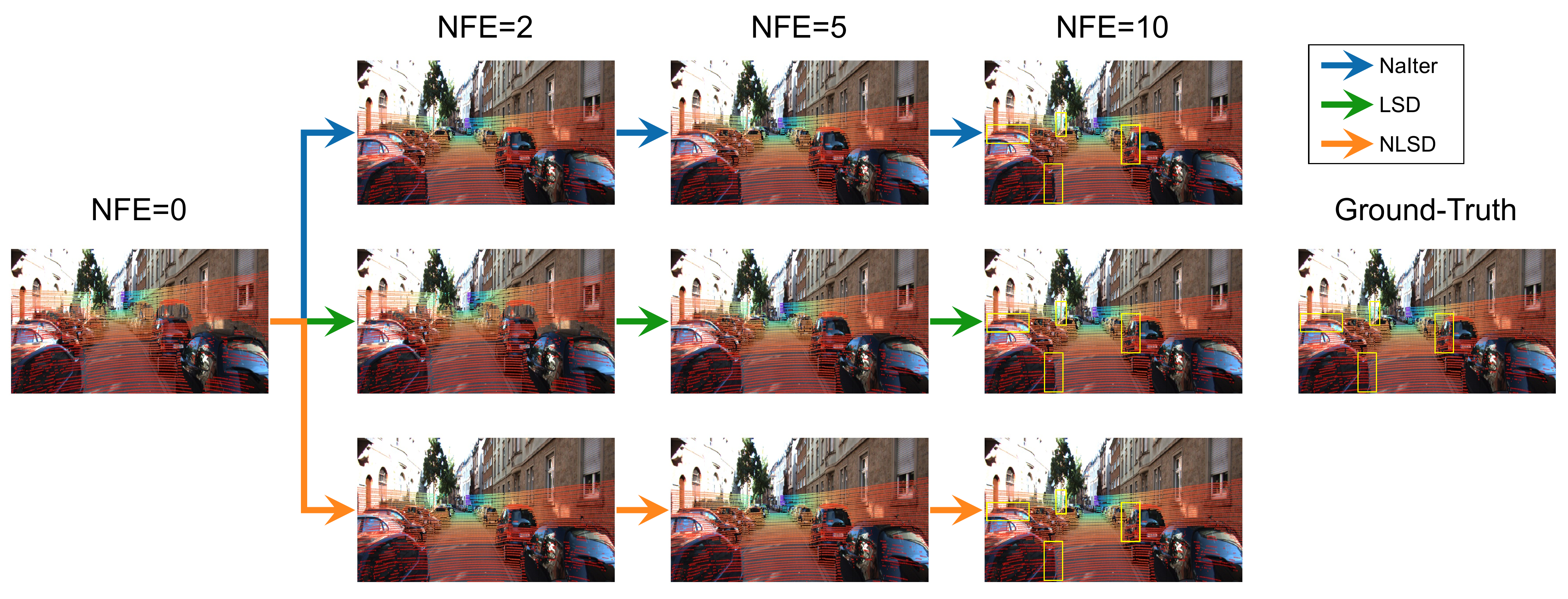}
    \caption{LiDAR projection maps of different iterative methods (from up to bottom: NaIter, LSD, NLSD). In addition to the initial state common to all three methods, we sampled three intermediate results at NFE=2, 5, and 10 over ten steps to facilitate comparison. Significant differences in their final states (NFE=10) are highlighted with yellow rectangles. The ground-truth calibrated state is also provided for reference.}
    \label{Fig.cmp_diff}
\end{figure*}
\begin{figure}[!t]
    \centering
    \includegraphics[width=0.98\linewidth]{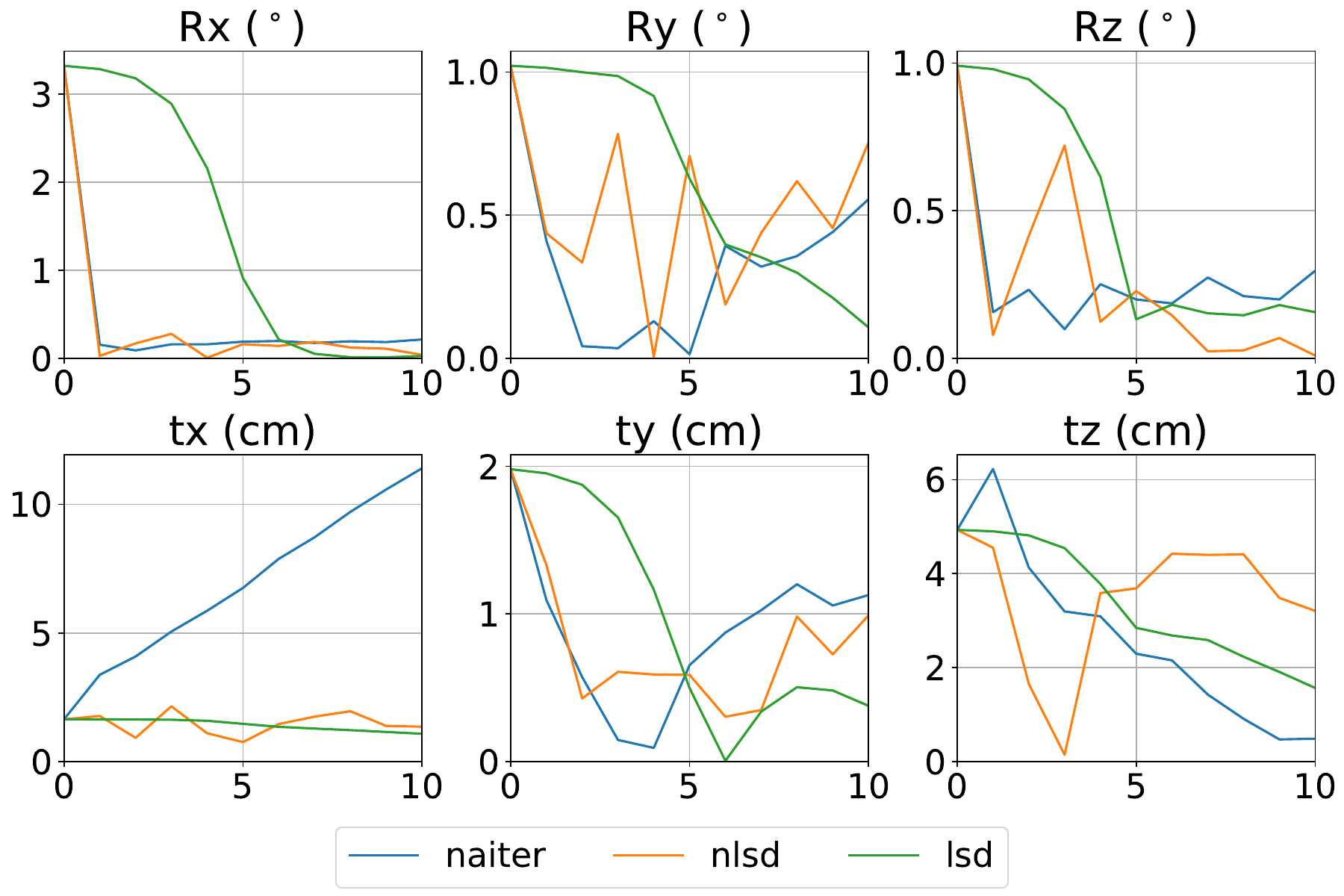}
    \caption{Error curves of different iterative methods \wrt an example scene. The x and y axes respectively denote NFE and the magnitude of error.}
    \label{Fig.err_curve}
\end{figure}
To qualitatively illustrate the differences in terms of iteration process among these methods, we draw LiDAR projection maps of an urban calibration scene over the course of the entire iterative calibration in \cref{Fig.cmp_diff}. Although NaIter and NLSD converge faster than LSD, the latter achieves superior final accuracy. The yellow rectangles in \cref{Fig.cmp_diff} indicate that several critical edges are better aligned using LSD compared to NLSD and NaIter. Furthermore, the corresponding error curves are plotted in \cref{Fig.err_curve}. The errors of six axes all basically decrease with the NFE using LSD, which is an advantage not observed with NLSD and NaIter.

From a theoretical perspective, Naiter simply calls $D_\theta(\cdot)$ repeatedly to refine the current extrinsic matrix. In contrast, both NLSD and LSD formulate the entire iterative calibration problem as a diffusion process where each correction step is treated as a single denoising step, leading to a more accurate and stable iterative process. The key differences between NLSD and LSD are listed as follows: first, NLSD defines the diffusion variable in the SE(3) space, whereas LSD does so in the $\mathfrak{se}(3)$ space; second, in generating $\bm{x}_t$, NLSD employs a combination of nonlinear perturbation and interpolation, while LSD relies solely on linear interpolation; third, their posterior distributions differ. Following the conventions in~\cite{SE3-Diffusion}, NLSD transforms both $\bm{H}_0$ and $\bm{H}_t$ into the $\mathfrak{se}(3)$ space for combinations, and then maps the result back to the SE(3) space to obtain $\bm{H}_{t-1}$, whereas LSD directly derives $\bm{x}_{t-1}$ through a linear combination of $\bm{x}_0$ and $\bm{x}_t$.  

We attribute the superior performance of LSD over NLSD to two main factors. First, due to the linearity of the diffusion variable, LSD's reverse process can be naturally formulated as an ODE process, leading to improved numerical accuracy—an advantage that is not applicable to NLSD because the computation of posterior $\bm{H}_{t-1}$ is nonlinear. Second, due to the linear interpolation in the $\mathfrak{se}(3)$ space, LSD avoids taking excessively large correction steps at the early iterations, thereby preserving room for further refinement if the initial prediction is insufficiently accurate.

\subsection{Efficiency Test}
We present the inference time per batch (with a batch size of 16) for each model in Table \ref{Table.efficiency}. All tests were conducted on a computer equipped with an NVIDIA RTX 4060 Laptop GPU and an Intel i7-12650H CPU. Since NaIter primarily involves repeated computations of $D_\theta$ with minimal additional operations, comparing the execution speed of the Single and NaIter models provides a fair assessment of the efficiency improvements achieved by our proposed buffering technique. Theoretically, NaIter's inference time should be at least ten times that of the single-step model; however, in practice, the real inference time is significantly shorter due to the buffering technique.  This technique reduces inference time by 21.35\% (LCCRAFT) to 51.15\% (CalibNet). Compared to NaIter, the implementation of LSD and NLSD introduces a moderate increase in computational time due to additional computations required by the noise scheduler. LSD incurs a slightly higher overhead due to the numerical approximation steps in the ODE solver.
\begin{table}[htbp]
\caption{Inference Time (ms) per Batch for Each Model}
    \label{Table.efficiency}
    \centering
    \begin{tabular}{crrrr}
    \toprule
    Method &Single$\downarrow$ &NaIter$\downarrow$ &NLSD$\downarrow$ &LSD$\downarrow$\\
    \midrule
    CalibNet~\cite{CalibNet} &40.67 &198.67 &226.01 &235.11\\
    RGGNet~\cite{RGGNet} &52.53 &321.16 &348.10 &356.91\\
    LCCNet~\cite{LCCNet} &65.36 &448.07 &475.99 &483.28\\
    LCCRAFT~\cite{LCCRAFT} &381.76 &3002.66 &3024.40 &3097.26\\
    \bottomrule
    \end{tabular}
\end{table}

\section{CONCLUSIONS}
\label{sec.conclusion}
In this study, we introduce a Linear Surrogate Diffusion (LSD) model for denoiser-agnostic iterative camera-LiDAR calibration. Experimental results indicate that LSD outperforms other baseline iterative methods in improving the surrogate model's accuracy and robustness and demonstrates the best stability. Efficiency tests confirm the effectiveness of our buffering technique. Our future research will focus on enhancing the iterative method's capacity to improve translation accuracy and on exploring specific geometric guidance for the proposed diffusion model.
\bibliographystyle{IEEEtran}
\bibliography{root}\ 


\end{document}